\documentclass[lettersize,journal]{IEEEtran}
\usepackage{amsmath,amsfonts}
\usepackage{algorithmic}
\usepackage{algorithm}
\usepackage{array}
\usepackage[caption=false,font=normalsize,labelfont=sf,textfont=sf]{subfig}
\usepackage{textcomp}
\usepackage{stfloats}
\usepackage{url}
\usepackage{verbatim}
\usepackage{graphicx}
\usepackage{cite}
\usepackage{acronym}
\usepackage{hyperref}
\usepackage{xurl}

\usepackage{acronym} 

\acrodef{SNN}[SNN]{Spiking Neural Network}
\acrodef{STDP}[STDP]{Spike-timing-dependent plasticity}
\acrodef{DG}[DG]{Dentate Gyrus}
\acrodef{PC}[PC]{Pyramidal Cell}
\acrodef{CA}[CA]{Cornu Ammonis}
\acrodef{LIF}[LIF]{Leaky Integrate-and-Fire}
\acrodef{EC}[EC]{Entorhinal Cortex}
\acrodef{ANN}[ANN]{Artificial Neural Network}
\acrodef{LIF}[LIF]{Leaky integrate-and-fire}
\acrodef{LTP}[LTP]{Long-Term Potentiation}
\acrodef{LTD}[LTD]{Long-Term Depression}
\acrodef{ANN}[ANN]{Artificial Neural Network}

\makeatletter
\def\subsubsection{%
  \@startsection
    {subsubsection}                 
    {3}                             
    {\z@}                    
    {3.5ex plus 1.5ex minus 1.5ex}  
    {0.7ex plus .5ex minus 0ex}     
    {\normalfont\normalsize\itshape}
}
\makeatother

\begin{document}

\title{A bio-inspired implementation of a sparse-learning spike-based hippocampus memory model}

\author{Daniel~Casanueva-Morato~\IEEEmembership{Member,~IEEE,}, Alvaro~Ayuso-Martinez~\IEEEmembership{Member,~IEEE,}, Juan~P.~Dominguez-Morales~\IEEEmembership{Member,~IEEE,}, Angel~Jimenez-Fernandez~\IEEEmembership{Member,~IEEE,}, Gabriel~Jimenez-Moreno~\IEEEmembership{Member,~IEEE}
\thanks{Manuscript received April 19, 2021; revised August 16, 2021. This research was partially supported by the Spanish grant MINDROB (PID2019-105556GB-C33/AEI/10.13039/501100011033). D. C.-M. was supported by a "Formaci\'{o}n de Personal Universitario" Scholarship from the Spanish Ministry of Education, Culture and Sport.}
\thanks{D.~C.-M., A.~A-M., J.~P.~D.-M., A.~J.-F. and G.~J.-M. are with the Robotics and Technology of Computers Lab, Universidad de Sevilla, Spain (email: dcasanueva@us.es).}}

\markboth{Journal of \LaTeX\ Class Files,~Vol.~14, No.~8, August~2021}%
{Shell \MakeLowercase{\textit{et al.}}: A Sample Article Using IEEEtran.cls for IEEE Journals}


\maketitle

\begin{abstract}

The nervous system, more specifically, the brain, is capable of solving complex problems simply and efficiently, far surpassing modern computers. In this regard, neuromorphic engineering is a research field that focuses on mimicking the basic principles that govern the brain in order to develop systems that achieve such computational capabilities. Within this field, bio-inspired learning and memory systems are still a challenge to be solved, and this is where the hippocampus is involved. It is the region of the brain that acts as a short-term memory, allowing the learning and unstructured and rapid storage of information from all the sensory nuclei of the cerebral cortex and its subsequent recall. In this work, we propose a novel bio-inspired memory model based on the hippocampus with the ability to learn memories, recall them from a cue (a part of the memory associated with the rest of the content) and even forget memories when trying to learn others with the same cue. This model has been implemented on the SpiNNaker hardware platform using Spiking Neural Networks, and a set of experiments and tests were performed to demonstrate its correct and expected operation. The proposed spike-based memory model generates spikes only when it receives an input, being energy efficient, and it needs 7 timesteps for the learning step and 6 timesteps for recalling a previously-stored memory. This work presents the first hardware implementation of a fully functional bio-inspired spike-based hippocampus memory model, paving the road for the development of future more complex neuromorphic systems.
\end{abstract}

\section{Introduction}
\label{sec:introduction}

\IEEEPARstart{T}{he} nervous system has demonstrated great computational capacity, adaptability and efficient resolution of complex problems \cite{budd2015early}. Neuromorphic engineering focuses on the study, design and implementation of hardware and software systems that mimic the basic structural and functional principles of the biological nervous system in order to achieve such energy-efficient computational capabilities observed in nature \cite{mead1990neuromorphic, indiveri2011neuromorphic}.

These systems are made up of artificial neurons interconnected by means of synapses and are responsible for the generation, processing and transmission of action potentials (asynchronous electrical pulses, also called spikes) that encode information. Learning in these neural networks is achieved through the plasticity of the synapses that connect the different neurons. This approach has great advantages in terms of energy consumption and real-time operation compared to traditional systems \cite{NeuromorphNature, zhu2020comprehensiveReview}. In order to manage these biological principles, a bio-inspired computational approach is needed. A specific type of biologically-plausible neural networks called \acp{SNN} is the most widely used for this purpose.

In this field, memory, learning and information storage systems are still a challenge to be solved, with much work to be done. In biology, the hippocampus is defined as a short-term memory responsible for the rapid storage of information from all sensory nuclei. This brain region is responsible for, among other high-level functions, collecting all the information from the environment to provide enough time for other nerve centres to process it, thus information that is frequently accessed will remain in memory for longer and information that is not frequently accessed will be forgotten~\cite{rolls2021brain}.


Some other works regarding this topic can be found in the literature, although these are scarce and most of them approach it from a theoretical perspective \cite{shiva2016continuous}. In the case of those that propose a computational model, in general, they do not offer sufficient information to reproduce them \cite{shiva2016continuous}, they have very low storage capacities \cite{tan2011associative}, \cite{tan2013hippocampus}, they are not purely bio-inspired approaches \cite{zhang2016hmsnn} or they present limitations regarding the information they can store \cite{tan2011associative}, \cite{tan2013hippocampus}, \cite{casanueva2022spike}.

Given these shortcomings, this paper proposes a bio-inspired spike-based computational memory model of the hippocampus implemented on the SpiNNaker hardware platform. This memory model is capable of learning and storing spiking information as a memory, as well as recalling it later by presenting the cue (part of the memory associated with the rest of the content) and even forgetting memories when trying to learn others with the same cue.

\subsection{Biological model of the hippocampus}
\label{subsec:biological_model}

The hippocampus or hippocampal system belongs to the limbic system, a region of the brain made up of a collection of brain structures involved in the control of emotional responses, smell, sleep and certain areas of the memory, among other functions. The limbic system is located within the cerebrum of the brain, underneath the cerebral cortex, next to the temporal cortex \cite{rajmohan2007limbic, roxo2011limbic}.

There is no universal agreement on the exact brain structures that make up the limbic system. However, one of the most widespread proposals, which is based on the functional aspect of the system, is given in \cite{rajmohan2007limbic}. According to this proposal, the limbic system consists mainly of: limbic lobe, hippocampal formation, amygdala, septal area and hypothalamus.

Focusing on the hippocampus, it is the brain structure that acts as an episodic memory for rapid and unstructured storage of information obtained from different areas of the neocortex. The neocortex is formed by the cerebral cortex and conducts very diverse and different information to the hippocampus; therefore, when indicating that the storage is unstructured, we are referring to the fact that the information is stored as an association of information regardless of its origin and semantic interpretation at a specific moment (or small interval) \cite{rolls2021brain}.

The events (association of information from different sensory sources) stored in the episodic memory lack interpretation or understanding; it is the semantic memory of the neocortex that gradually builds and adjusts, on the basis of much accumulated information, the semantic representation. In the semantic memory, the stored information is structured and a higher level of abstraction is given. Thus, both memory systems are complementary; the episodic events that are actively recalled will be used in the neocortical representation and, thus, in the semantic memory, while all the episodic memories that are not recalled will be gradually replaced (forgotten) by others in the hippocampal episodic memory \cite{rolls2021brain, anand2012hippocampus}.


This functionality is achieved by means of the three layers that form the hippocampus, as is shown in Fig.~\ref{fig:hippstructdetailed}:


\begin{figure}[htbp]
    \centerline{\includegraphics[width=0.35\textwidth]{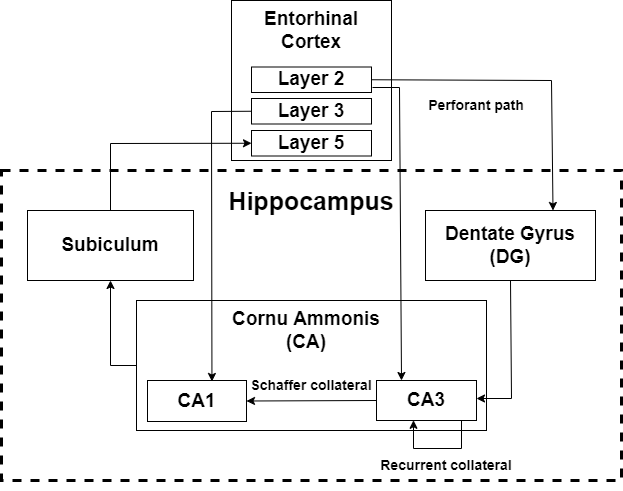}}
    \caption{Block diagram of the biological architecture of the hippocampus with its layers and detailed connecting pathways between them.}
    \label{fig:hippstructdetailed}
\end{figure}

\begin{figure}[htbp]
    \centering
    \centerline{\includegraphics[width=0.35\textwidth]{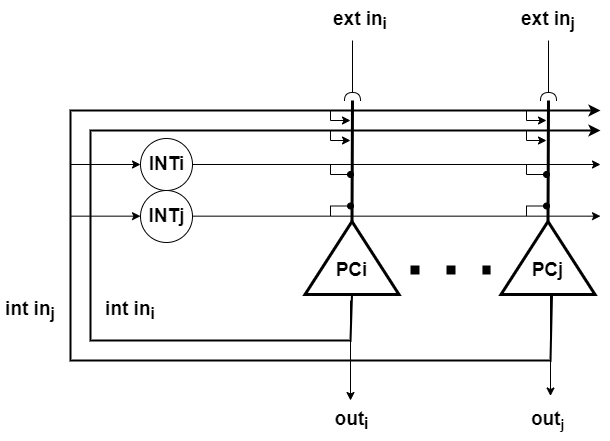}}
    \caption{Internal structure of \ac{CA}3. It consists of a recurrent collateral network of \acp{PC} with external inputs from \ac{DG} (ext) and internal inputs from the collateral connections (int), as well as collateral inhibitory inputs from a population of interneurons (INT) that regulate the rate of activity of the network. All the connections are excitatory (lines ending with an arrowhead) except for the projections between the interneurons back to the input of the \acp{PC}, which are inhibitory (lines ending with a circle).}
    \label{fig:ca3struct}
\end{figure}

\begin{itemize}
    \item \textbf{\acf{DG}}: this is the input region of information from \ac{EC} and it plays an important role in pattern separation by spreading the information content. This function allows reducing the degree of correlation between different input patterns. Therefore, the aim of this region is to increase the storage capacity and improve the access to the stored information, as this information is more orthogonal and thus easier to identify unambiguously \cite{anand2012hippocampus}.
    
    \item \textbf{Proper hippocampus (\acf{CA})}: it consists of \acp{PC}, and it is divided into \ac{CA}1, \ac{CA}2, \ac{CA}3 and \ac{CA}4, with \ac{CA}1 and \ac{CA}3 being the most relevant in terms of currently known functionality. It is where the input information is stored and the stored information is recalled and recoded, when requested from the outside \cite{wible2013hippocampal}.
    
    \begin{itemize}
        \item \textbf{\ac{CA}3}: it is a recurrent collateral network of \ac{PC} neurons with an oscillating activity regulated by recurrent inhibitory interneurons, as can be seen in Fig.~\ref{fig:ca3struct}. In terms of its architecture, it is an autoassociative or attractor memory, to which the capacity to store all the input information quickly and unstructured is added, thereby making it short term memory. Taking into account that the input information comes from all the sensory flows of the brain, the hippocampal memory can be considered to act as an episodic memory \cite{rolls2021brain}.
        
        At the level of collateral synapses, the neurons forming \ac{CA}3 are connected to a set of other \ac{CA}3 neurons, and some even with themselves (to a lesser extent). The key for storing information in this type of network lies in the set of dynamic synapses and the distributions of synapses that exist between the different neurons \cite{rolls2021brain}.
        
        \item \textbf{\ac{CA}1}: it presents a competitive network architecture to recode the input information. It receives the memories that are wanted to be recalled from \ac{CA}3 and it is in charge of performing the necessary recoding (the inverse of that achieved in \ac{DG}) that allows for the total recovery of the original memory~\cite{rolls2021brain}.
    \end{itemize}
    
    \item \textbf{Subiculum}: it receives the output from the hippocampus and redirects it back to \ac{EC}.
\end{itemize}

As in the case of the limbic system, there is no consensus on which layers are part of the hippocampal system  \cite{anand2012hippocampus, wible2013hippocampal}. The presented biological model and its structure was proposed in \cite{rolls2021brain}, and it is the most widely used in the literature. It is also consistent with the model of the limbic system presented, as it considers \ac{EC} as a component that is external to the hippocampus, specifically belonging to the limbic lobe.

Although it is not part of the hippocampal system, the \ac{EC} (Brodmann's area 28) serves as its main information input and output pathway. It has as many input (forward) as output (backward) projections. Through this cortex, the hippocampus receives information from all the different sensory information streams in the brain (including spatial) coming from the neocortex and sends the processed information back to these sensory stream units \cite{rolls2021brain} \cite{anand2012hippocampus}.

Therefore, the hippocampal learning process starts by receiving the input of all sensory information from \ac{EC} through \ac{DG}. In \ac{DG}, the information is dispersed to be as different as possible from other previous or subsequent similar inputs and then passes to \ac{CA}3, where this information is stored as a memory. In order to recall this memory, the only thing needed is to present a part of it (recall cue) to \ac{CA}3, which starts a process of obtaining the complete memory called completion. After completion, the complete memory passes to \ac{CA}1, where it is recoded to recover its original information, preparing it for its return to the neocortex via the subiculum (through \ac{EC}), where it would be used \cite{rolls2021brain}.

\subsection{Related work}
\label{subsec:related_work}


Designing and implementing different bio-inspired memory models has already been explored and addressed in the literature. Works such as \cite{tan2011associative} and \cite{tan2013hippocampus} propose a model with an architecture that is biologically inspired in the hippocampus and that consists of a double \ac{CA}3 network. The first \ac{CA}3 network is responsible for the storage of input information and the second network allows for the recalling of the sequence based on a cue (the first neuron in the sequence). This memory is capable of storing a sequence of up to 11 elements, with each element being the activation of an input neuron. In other words, it has the capacity to store the activation of 11 neurons independently in a 1-by-1 sequence. Other authors, such as \cite{zhang2016hmsnn}, proposed a model of the hippocampus formed by \ac{DG} and \ac{CA}3; however, it is not purely spiking or bio-inspired, since it relies on functions commonly used in \acp{ANN} to model certain parts of the memory.

In \cite{casanueva2022spike}, two oscillating memory models based on the hippocampus with different degrees of biological abstraction are proposed: one is closer to biology and the other one is more functional. Although both models are capable of storing and recalling different memories, they have certain limitations, such as the difficulty of storing non-orthogonal memories (memories whose activity contains activations of neurons in common), and combining learning and recalling operations. Limitations also found in the associative memory model proposed in \cite{he2019constructing}, where an intermediate phase of synapse pruning is added in exchange for sacrificing the dynamism of the synapses.


Other references can be found in the literature regarding this topic, focusing on a more theoretical approach, such as \cite{shiva2016continuous}, in which a \ac{CA}3 model for short-term storage of information is mathematically proposed. However, no details are given regarding its operation or characteristics.

In \cite{bush2010dual}, the plasticity of dynamic connections in the hippocampus is studied, specifically in \ac{CA}3. The mentioned study also proposed a learning rule that adds the theta signal to the \ac{STDP} mechanism (see Section~\ref{sec:materials_and_methods}), which characterises the oscillatory activity of this brain region. This rule is a bio-inspired alternative capable of achieving time and frequency coding of the input activity patterns. Making use of this rule, the authors demonstrated its functionality with an autoassociative network model; however, the paper focuses on the practical demonstration of the presented theoretical model of a new bio-inspired learning law. Other works, such as \cite{oess2017computational}, proposed a bio-inspired spatial navigation mechanism in which a \ac{CA}3 model is used. However, the \ac{CA}3 model used in this proof-of-concept has no learning mechanism, and it is purely static with previously stored information.


\subsection{Main contributions}
\label{subsec:main_contributions}

The main contributions of this work include the following:
\begin{itemize}
    \item A fully-functional spike-based bio-inspired hippocampal memory model is proposed. It is capable of storing, recalling and forgetting both orthogonal and non-orthogonal patterns. In addition, it presents a mechanism to forget non-relevant information.
    \item The proposed model was simulated in software and also implemented and emulated on the SpiNNaker hardware platform.
    \item The architecture of the model is parameterised, allowing the user to define its capacity, including the size of each memory and the number of memories that can be stored at the same time.
    \item The proposed model is evaluated by means of a wide variety of tests and benchmarks to prove the correct function of the model itself and all the implemented functionalities in detail. 
    \item The source code is publicly available, together with the documentation including all the necessary details regarding the \ac{SNN} architectures.
\end{itemize}


The rest of the paper is structured as follows: Section~\ref{sec:materials_and_methods} presents the materials and methods used in this work. In Section~\ref{sec:sparse_learning_hippocampus_model}, the proposed spike-based computational model of the hippocampus is detailed, including its architecture (Section~\ref{subsec:hippocampus_architecture}), its operating principle and implemented functionalities (Section~\ref{subsec:operating_principle}), and the resources needed together with the timing constraints (Section~\ref{subsec:hippocampus_resources}). The experiments performed to evaluate the functionality and performance of the proposed model are explained in Section~\ref{sec:experiments_and_results}, along with the results obtained. Then, in Section~\ref{sec:discussion}, the results of the experiments are discussed. Finally, the conclusions of the paper are presented in Section~\ref{sec:conclusions}.

\section{Materials and methods}
\label{sec:materials_and_methods}

\subsection{Spiking Neural Networks}
\label{subsec:spiking_neural_networks}



The third generation of neural networks, \acp{SNN}, is one of the best options when working with a bio-inspired computational memory model of the hippocampus. These allow allow creating networks of neurons from components (neurons, synapses and learning rules) that are inspired in biology and aim to mimic it in order to incorporate the neurocomputational capabilities found in nature \cite{ahmed2020brain}.

The operation and communication of \acp{SNN} are based on spikes, which are asynchronous events that incorporate the concept of space and time into neural networks through connectivity and plasticity. These networks are very efficient from a computational point of view, as they only operate when an event occurs (when input spikes are received), particularly, only the components of the network that are affected at that moment, due to their asynchronous behaviour. Furthermore, the learning process is distributed throughout the network thanks to the \ac{STDP} learning mechanism \cite{caporale2008spike}, which adapts the weight of the synapses between local neurons based on the relative timing information between pre- and post-synaptic spikes \cite{ahmed2020brain}. All these aforementioned aspects make it possible to interpret and understand \ac{SNN} models in detail, enabling the extraction of the set of rules that govern them \cite{lobo2020spiking}.


\acp{SNN} present a set of characteristics and features that favour their implementation in hardware platforms in comparison to traditional neural networks, allowing for greater computational efficiency with lower power consumption: multiplications are replaced with adders and shifts, the information that is transmitted between neurons is only 1-bit size instead of integer or floating point numbers, etc. \cite{lobo2020spiking}.

At the biological level, there is a wide variety of models for each of the basic components that are present in a neural network (neurons, synapses and learning mechanisms) depending on the level of abstraction and, therefore, the level of neurocomputational complexity to be achieved.

Regarding neuron models, \ac{LIF} is the most widely used in the literature, as it is the most computationally simple, and thus it allows implementing of models with the largest number of neuronal circuits \cite{tavanaei2019deep}. The standard \ac{LIF} model, initially presented in \cite{stein1965theoretical}, describes the behaviour of a neuron by means of an RC circuit. The neuron has a variable membrane potential equivalent to the voltage measurement in parallel across the resistor and capacitor. With no input or output activity, the membrane potential of the neuron will remain stable at a resting value, which will be perturbed by the arrival of a  spike from another neuron. This current will have a positive or negative influence on the membrane potential, which, if it reaches a set threshold, will make the neuron generate a spike. After this, its membrane potential drops to a reset value, slightly below the resting value, and will stop integrating new input activity for a period of time called refractory period. This mechanism is known as integrate-and-fire. By default, when the membrane potential is modified, it tends to return to its resting value, which results in a constant decrease at each time instant, called leak.

Synapses are modelled as weighted edges in a graph linking a source neuron (presynaptic neuron) to a target neuron (postsynaptic neuron). The weight represents the influence that the activity of the presynaptic neuron exerts on the activity of the postsynaptic neuron when the former generates spikes. This influence is translated into the amount of increase or decrease in potential that a presynaptic spike wields on the postsynaptic neuron. The learning mechanisms of this type of neuronal network focus on the plasticity of the synapses, i.e., the ability to create, eliminate or modify the weight of the synapses dynamically.

Among the different learning mechanisms based on synaptic plasticity, the most prominent and widely used is \ac{STDP}. It is based on a Hebbian learning algorithm in which the weight of synapses are modified in proportion to the degree of temporal correlation between the activity of pre- and post-synaptic neurons \cite{sjostrom2010spike}. The weight of the synapse is increased if a presynaptic spike occurs before a postsynaptic spike (\ac{LTP}), and decreased (\ac{LTD}) otherwise, by an amount proportional to the time difference between the two spikes (the closer in time, the greater the weight change).

Different hardware platforms particularly designed for implementing and simulating \acp{SNN} can be found in the literature. Some of the most well-known hardware platforms are SpiNNaker \cite{furber2014spinnaker}, Loihi \cite{davies2018loihi} and TrueNorth \cite{merolla2014million}. In this work, we used SpiNNaker as the hardware platform in which the different \ac{SNN} models presented were implemented and emulated.

\subsection{SpiNNaker}
\label{subsec:spinnaker}

SpiNNaker \cite{furber2014spinnaker} is a massively-parallel multi-core computing system, which was designed to allow modeling very large \acp{SNN} in real time. Each SpiNNaker chip consists of 18 general-purpose ARM968 cores, running at 200 MHz, which communicate the information by means of packets carried by a custom interconnect fabric. In this work, a SpiNN-5 machine was used, which consists of 48 SpiNNaker chips in total, and, therefore, has 864 ARM processor cores (768 application cores, 48 Monitor Processors and 48 spare cores). A 100 Mbps Ethernet connection is used as an I/O interface and for sending scripts and commands to the board. A custom software package called sPyNNaker \cite{rhodes2018spynnaker} allows running PyNN \cite{davison2009pynn} simulations directly on the SpiNNaker board, making the platform very straight-forward to work with, since all the codes regarding the design and implementation of \acp{SNN} can be done using high-level functions described in Python programming language.

In terms of temporal characteristics, SpiNNaker works with a time step of 1~ms as the minimum unit of time. Therefore, the design of any model on this platform considers instants of time with a resolution of that timestep. \ac{STDP} is implemented in SpiNNaker, although it presents some particularities: the evaluation of the weight change (sum of cumulative increments and decrements) in a synapse is produced only on presynaptic spikes \cite{jin2010implementing}, which has to be taken into account when using that learning rule.

\section{Sparse-learning hippocampus computational memory model}
\label{sec:sparse_learning_hippocampus_model}

This paper proposes an implementation of a novel spike-based computational memory model inspired on the hippocampus, whose main architecture can be seen in Fig.~\ref{fig:memoryarch}\footnote{From now on, synapses from population A to B will be referred to as A-B for the sake of simplicity.}. The architecture and information flow is very similar to that present in the biological model of the hippocampus (see Section~\ref{subsec:biological_model}), where the information that is received from \ac{EC} (INPUT) passes through \ac{DG}, \ac{CA}3, \ac{CA}1 and, finally, the OUTPUT returns back to \ac{EC} (see Fig.~\ref{fig:hippstructdetailed}).

The proposed memory model presents an architecture whose size depends on two parameters:

\begin{itemize}
    \item \textbf{Maximum number of cues (cueSize)}: the maximum number of memories that can be stored at the same time.
    
    \item \textbf{Content size (contSize)}: the size of the content of the memories that can be stored in terms of neuron activity, i.e., the number of neurons dedicated to the storage of memories minus the cue information.
\end{itemize}

From these two variables, the sizes of the different layers and input and output activity arrays can be calculated automatically, as well as the interconnection configuration of the different internal blocks of the memory.

The broad functioning of the model is that of a memory with the capacity to learn and recall memories following a hippocampal-based workflow. In order for the memories to be stored during the learning process and retrieved on the recalling step, they must contain a cue. Therefore, the input (and output) array can be divided into two main parts: a set of neurons that contain the cue and the remaining neurons that refer to the content of the memory itself.

The learning process requires not only the storage of the memory, but also the association of the memory with the cue that identifies it. On the other hand, the process of recalling a memory only requires that the activity related to the cue of the memory is introduced in the input and, internally, the memory is responsible for retrieving the entire memory. The model also presents a forgetting mechanism with which, when a new memory whose cue is the same as that used by another memory is learned, the model is able to forget the old memory and store the new one.

There is no prior interpretation of the content of memories. Since they depend completely on their source, a memory can be formed by the association of information from different sources. This information has a spatio-temporal coding, thus the concept of memory refers to the concatenation of activity of the different input and/or output neurons at a specific moment. Therefore, the activity of different neurons at the same instant corresponds to the same memory, and the activity of the same neuron at different instants refers to different memories.

\subsection{Architecture}
\label{subsec:hippocampus_architecture}

\begin{figure}[!t]
    \centering
    \includegraphics[width=0.30\textwidth]{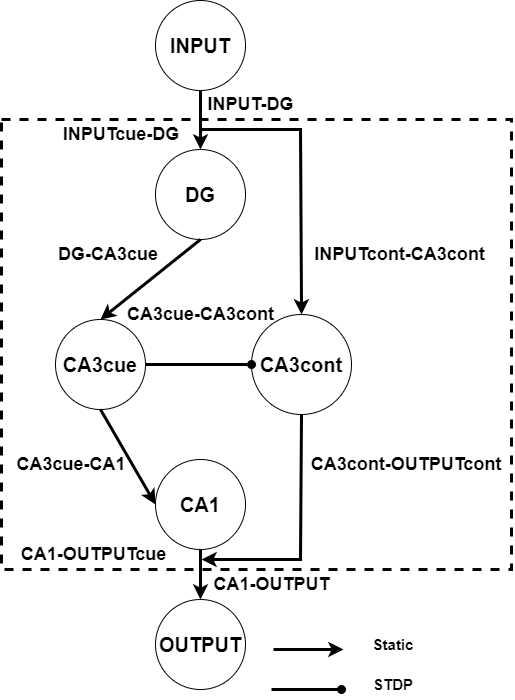}
    \caption{Architecture of the hippocampus computational memory model proposed. The blocks that correspond to the architecture are framed with a dashed line, which consists of \ac{DG}, \ac{CA}3cue, \ac{CA}3cont and \ac{CA}1. All interconnections are excitatory except \ac{CA}3cue-\ac{CA}3cont, which have dynamic synapses using \ac{STDP}.}
    \label{fig:memoryarch}
\end{figure}

Fig.~\ref{fig:memoryarch} shows the architecture of the proposed memory model, which is divided into 3 main layers, with their respective interconnections: \ac{DG}, \ac{CA}3 and \ac{CA}1.

\begin{table*}[]
\centering
\caption{Brief explanation of the parameters needed to define a standard \ac{LIF} neuron and values used for each parameter to define the behaviour of neurons in different components of the proposed hippocampus computational memory model.}
\label{tab:neuronparams}
\resizebox{\textwidth}{!}{%
\begin{tabular}{|l|l|c|c|c|}
\hline
\textbf{Neuron parameters} & \textbf{Overview}                                                                               & \multicolumn{1}{l|}{\textbf{DG and CA1}} & \multicolumn{1}{l|}{\textbf{CA3 cue}} & \multicolumn{1}{l|}{\textbf{CA3 cont}} \\ \hline
$c_{m}$ (nF)               & Capacitance                                                                                     & 0.1                                      & 0.27                                  & 0.27                                  \\ \hline
$\tau_{m}$ (ms)            & Time-constant of the RC circuit                                                                 & 0.1                                      & 3.0                                   & 3.0                                   \\ \hline
$\tau_{refrac}$ (ms)       & Time that the neuron is disabled after generating a spike and stops integrating any input spike & 0.0                                      & 1.0                                   & 1.0                                   \\ \hline
$\tau_{synExc}$ (ms)       & Excitatory input current decay time-constants                                                   & 0.1                                      & 0.3                                   & 0.3                                   \\ \hline
$\tau_{synInh}$ (ms)       & Inhibitory input current decay time-constants                                                   & 0.1                                      & 0.3                                   & 0.3                                   \\ \hline
$v_{reset}$ (mV)           & Membrane potentials at which the neuron is set immediately after generating a spike             & -65.0                                    & -60.0                                 & -60.0                                 \\ \hline
$v_{rest}$ (mV)            & Membrane potential to which the neuron tends when the membrane potential is altered.            & -65.0                                    & -60.0                                 & -60.0                                 \\ \hline
$v_{thresh}$ (mV)          & Threshold at which the neuron spikes                                                            & -64.91                                   & -57.0                                 & -57.5                                 \\ \hline
\end{tabular}%
}
\end{table*}

\begin{table*}[]
\centering
\caption{Brief explanation of the parameters defining static and dynamic synapses (with STDP), and the values used for each of them throughout the hippocampus computational memory model. Values that do not apply to the type of synapse described are marked with "-".}
\label{tab:synparams}
\resizebox{\textwidth}{!}{%
\begin{tabular}{|l|l|c|c|c|c|c|}
\hline
\textbf{Synapses parameters} & \textbf{Overview}                                                           & \multicolumn{1}{l|}{\textbf{CA3cue-CA3cont}} & \multicolumn{1}{l|}{\textbf{\begin{tabular}[c]{@{}l@{}}IL-DG\\ CA3cue-CA1\end{tabular}}} & \multicolumn{1}{l|}{\textbf{\begin{tabular}[c]{@{}l@{}}DG-CA3cue\\ CA1-OL\end{tabular}}} & \multicolumn{1}{l|}{\textbf{IL-CA3cont}} & \multicolumn{1}{l|}{\textbf{CA3cont-OL}} \\ \hline
receptor type                & Type of synapse                                                             & STDP                                        & excitatory                                                                               & excitatory                                                                               & excitatory                              & excitatory                              \\ \hline
init weight (nA)             & Initial weight of the synapse                                               & 0.0                                         & 1.0                                                                                      & 6.0                                                                                      & 6.0                                     & 6.0                                     \\ \hline
delay (ms)                   & Time taken for the presynaptic spike to reach the postsynaptic neuron       & 1.0                                         & 1.0                                                                                      & 1.0                                                                                      & 4.0                                     & 2.0                                     \\ \hline
$\tau_{plus}$ (ms)           & Decay time-constants that control the amount of weight increase (only STDP) & 3.0                                         & -                                                                                        & -                                                                                        & -                                       & -                                       \\ \hline
$\tau_{minus}$ (ms)          & Decay time-constants that control the amount of weight decrease (only STDP) & 3.0                                         & -                                                                                        & -                                                                                        & -                                       & -                                       \\ \hline
$A_{plus}$                   & Maximum weight to add during potentiation (only STDP)                       & 6.0                                         & -                                                                                        & -                                                                                        & -                                       & -                                       \\ \hline
$A_{minus}$                  & Maximum weight to subtract during depression (only STDP)                    & 6.0                                         & -                                                                                        & -                                                                                        & -                                       & -                                       \\ \hline
$w_{max}$ (nA)               & Maximum possible weight of the synapse (only STDP)                          & 6.0                                         & -                                                                                        & -                                                                                        & -                                       & -                                       \\ \hline
$w_{min}$ (nA)               & Minimum possible weight of the synapse (only STDP)                          & 0.0                                         & -                                                                                        & -                                                                                        & -                                       & -                                       \\ \hline
\end{tabular}%
}
\end{table*}

\subsubsection{Dentate gyrus}
\label{subsubsec:dentate_gyrus}
It is the input layer and is responsible for recoding the memories that reach the CA3 layer in order to increase their dispersion, that is, to achieve a greater degree of differentiation between the different memories stored. At the biological level, it is still unknown how the dispersion carried out by \ac{DG} is performed and whether it is applied homogeneously or not, but it is known that it achieves a high degree of dispersion \cite{rolls2021brain}. Thus, for the proposed model, we considered the maximum possible expression of dispersion: the one-hot representation of the content of memories. 


One-hot encoding in the spiking domain refers to the case where, for each combination of activation of different input neurons, there is an output neuron that is uniquely activated, identifying that input combination. However, this recoding of the input memory will not be applied homogeneously over the entire input, but will be restricted only to the part of it corresponding to the cue. This is due to the fact that, as was previously mentioned, storing and recalling the memory depends mainly on the part corresponding to the cue. Therefore, the dispersion should only affect this part of the input. This way, only the right and necessary information that forms the memory is dispersed, which does not only optimise the storage of the memory, but it also simplifies the reverse decoding process carried out in \ac{CA}1. 

Thus, the proposed \ac{DG} model is responsible for the heterogeneous recoding of the input memory, converting the information carried by the neurons corresponding to the memory cue into a one-hot representation, regardless of the type of operation (learning, recalling or forgetting) performed. This function is carried out by means of a spiking decoder, proposed in \cite{alvaroMemoria2022}. It is a neuronal circuit based on spike-based logic with a functionality and topological structure similar to that of a digital decoder. To achieve this, a set of spiking building blocks are used, which are equivalent to digital logic gates, although in the spiking domain \cite{AlvaroIJCNN}.

\subsubsection{CA3}
\label{subsubsec:ca3}

The proposed model consists of a recurrent collateral network whose synapses are unidirectional connections only between the neurons that encode the part of the cue towards the neurons that encode the rest of the content of the memory. These synapses are created dynamically following a learning rule based on the \ac{STDP} mechanism \cite{jin2010implementing}. 

Therefore, \ac{CA}3 would consist of two subpopulations of neurons, one representing the part corresponding to the cue and, the other, the remaining content of the memory, connected to dynamic synapses based on the aforementioned learning rule. The first subpopulation is called Ca3cue and the second CA3cont, as can be seen in Fig.~\ref{fig:memoryarch}.

If the synapses that model the connections were static, we would be dealing with fixed associations generated manually (hardcoded). By having dynamic synapses, it is ensured that the content to be stored is also dynamic, being able to associate different memories with each cue that may change over time. This dynamism has been achieved with a \ac{STDP} rule, whose parameters are shown in Tab.~\ref{tab:synparams}. This configuration of the parameters allows for:

\begin{itemize}
    \item A fast stabilisation of the dynamic synapses and, thus, of the network, after performing any operation on memory (mainly learning and forgetting). This is achieved with low values of $\tau_{plus}$ and $\tau_{minus}$, as it (indirectly) reduces the time spent by the synapses to integrate spikes in order to modify their plasticity.
    
    \item Rapid changes in synaptic weights when performing operations that affect the plasticity of synapses. This is a consequence of using symmetrical maximum values for both positive and negative reinforcement of the synaptic weights ($A_{plus}$ and $A_{minus}$), equal to the maximum possible value they can adopt ($w_{max}$). In this way, each time there is a positive change, due to learning, or a negative change, due to forgetting, it will be large enough to lower or raise the weight to the maximum or minimum value by only presenting a few spike repetitions (which form the memory) to the network. Therefore, the amount of iterations needed for the same memory to be inputted in order to be learned (and forget the previous one at the same time) is reduced.
\end{itemize} 

The \ac{CA}3 module receives the memory encoded by \ac{DG} as input: the one-hot representation of the cue arrives at Ca3cue, and the rest of the information of the memory arrives at CA3cont without any encoding with respect to the original one. When a complete memory is inputted, the activity arrives at the same time to those two populations and the \ac{STDP} rule is activated, learning the new memory and forgetting the old one that had the same cue. In the case that only the cue is inputted, the rest of the memory associated with that cue will be evoked. In both cases, the output of this network is partially recoded and needs a reverse process to recover its original format. Thus, the output of \ac{CA}3 (specifically \emph{Ca3cue}) is connected to \ac{CA}1, and \emph{CA3cont} is passed directly to the memory output.

\subsubsection{CA1}
\label{subsubsec:ca1}

This is the output layer of the hippocampal memory and its goal is to re-encode the memory to recover the original input format. At a functional level, \ac{CA}1 performs the inverse operation of that carried out in \ac{DG} to eliminate the degree of dispersion added to the memory, which was necessary for its storage. As is indicated in the previous layers, the recoding is heterogeneous and is applied only to the information corresponding to the cue. Therefore, \ac{CA}1 receives only the output of \emph{Ca3cue}, as the activity of the remaining neurons encoding the rest of the memory maintains the original format. 


\ac{CA}1 was implemented using a spike-based encoder, which follows the same basis of operation and design as the spike-based decoder used in \ac{DG} \cite{alvaroMemoria2022}. It is a \ac{SNN} with the same functionality and topology as a digital encoder, although built with spiking building blocks \cite{AlvaroIJCNN}.


Thanks to this component, the original format of the cue can be retrieved and, with this, the proposed model is able to output the original and complete memory after both the recalling and learning operation, which is then ready to be interpreted or used in other external modules or as feedback to the memory itself.

\subsubsection{Interconnections}
\label{subsubsec:connections}

The proposed memory model is made up of 5 neuronal layers. These neuronal layers have a set of connections between them, whose defining synapse parameters are shown in Tab.~\ref{tab:synparams}.


We start with a layer of input neurons to the memory from the outside, whose activity will contain either the complete memory to be stored or only the necessary cue for its recall. The activity related to the cue must undergo a recoding that increases its dispersion; therefore, it has to pass through DG before reaching CA3 (INPUT-DG). Once the cue has been recoded, it will pass to the CA3 subpopulation, CA3cue (DG-CA3cue), just at the same instant that the remaining part of the memory arrives at the CA3cont subpopulation (INPUT-CA3cont). Thus, the INPUT-CA3cont synapses are delayed long enough to allow the cue to be encoded. 

In CA3, synapses that go from CA3cue to CA3cont (CA3cue-CA3cont) can be found, which follow a STDP learning rule. These are the connections that store the memories in memory, so they must be dynamic. Regardless of the operation, CA3 activity will be passed to the memory output; however, the cue has to be re-encoded back to its original format via CA1 (CA3cue-CA1). Once re-encoded, the cue will be outputted from CA1 (CA1-OUTPUT), at the same time as the content arrives from CA3cont (CA3cont-OUTPUT). For this to happen, the latter synapses must have a long enough delay to allow CA1 to re-encode the cue, so that they arrive at the same time.

All of these static synapses are excitatory, have a weight such that the activation of a presynaptic neuron generates a spike on its postsynaptic neuron, and maintain a 1-to-1 connection ratio, i.e., a presynaptic neuron connects to only one postsynaptic neuron. The only connections that are an exception are the dynamic synapses that govern the CA3 recurrent network. These are all-to-all connections from one subpopulation to the next and their weight, as the name suggests, is dynamic and varies throughout the run following the STDP algorithm in order to store memories.

In terms of the number of synapses connected between layers, the following can be found:
\begin{itemize}
    \item INPUT-DG: the number of neurons required to encode the maximum number of cues, i.e., $\lceil\log_{2}(cueSize)\rceil$
    \item DG-CA3cue: $cueSize$
    \item INPUT-CA3cont: $contSize$
    \item CA3cue-CA3cont: $cueSize * contSize$
    \item CA3cue-CA1: $cueSize$
    \item CA1-OUTPUT: $\lceil\log_{2}(cueSize)\rceil$
    \item CA3cont-OUTPUT: $contSize$
\end{itemize}

\subsection{Operating principle}
\label{subsec:operating_principle}

\subsubsection{Input and output}
\label{subsubsec:input_and_output}

The input and output populations of the proposed model have exactly the same number of neurons, which depends on \textit{cueSize} and \textit{contSize} (see Section~\ref{sec:sparse_learning_hippocampus_model}). The input array contains the part corresponding to the cue in its lowest positions, and the  rest corresponds to the content of the memory. Specifically, its size is defined by equation \ref{eq:inputsize}.

\begin{equation}
\label{eq:inputsize}
input size = \lceil\log_{2}(cueSize)\rceil + contSize
\end{equation}

When indicating the maximum number of cues that can be stored in memory at the same time, we refer, indirectly, to the number of possible combinations of the activations of the input/output neurons that are used for the cue. Therefore, following the calculation of the conversion from binary neuron activity to one-hot representation achieved in \ac{DG}, the number of neurons used as a cue can be calculated through the first summand in equation~\ref{eq:inputsize}. The second term of the equation refers to the remaining part of the memory that remains unchanged during internal processing.

\subsubsection{Implemented functionality}
\label{subsubsec:implemented_functionality}
The proposed model was designed to perform a set of memory-related operations, including:

\begin{itemize}
    \item Learning: the learning of a memory (write operation) is performed by holding the entire memory sample for 3 consecutive timesteps (3~ms) at the input. The memory must have the structure indicated in Section~\ref{subsubsec:input_and_output}. The cue part will be re-encoded and arrive at \ac{CA}3 at the same instant as the remaining part of the memory. This way, the \ac{STDP} rule will be activated at the CA3cue-CA3cont synapses that connect the neurons with active cues to the neurons with the rest of the memory content that are also active.
    
    \item Recall: to recall a memory (read operation), it is only necessary to input the cue for a single timestep (1~ms). The recoded cue produces a spike in the corresponding CA3cue neuron and, in turn, this makes the set of CA3cont neurons whose synapses have a reinforced weight spike as a consequence of the previous learning performed. In other words, only those CA3cont neurons that correspond to the learned memory associated with that cue will spike.
    
    \item Forgetting: the learning operation is not affected by the previous content of the memory, i.e., the learning of the memory is achieved regardless of whether or not something was previously learned using the same cue. This is because, indirectly, the forgetting mechanism is implemented within the learning mechanism. When a cue arrives, the neurons that stored the rest of the content of the memory are activated in CA3cont. However, if that cue is inputted together with the rest of the content of another memory (a learning operation using a cue already used for another memory), the recall will occur right in the middle of the activations corresponding to the learning (see Section~\ref{sec:experiments_and_results}). Consequently, the weight of the synapses that connect with those CA3cont neurons of the new memory will increase and the weight of the synapses with those corresponding to the old memory will decrease by the same amount. In short, the previous memory is forgotten in order to store the new one, due to the activation of those neurons of the old memory that is forgotten during the learning of the new one.
\end{itemize}

During these operations, the input activity passes through the different layers of the model. The processing times required in each of the different parts of the neural network are: 1 ms for INPUT-DG, 2 ms for DG, 1 ms for DG-CA3, 1 ms for CA3, 1 ms for CA3-CA1, 1 ms for CA1, 1 ms for CA1-OUTPUT and 4 ms for CA3 inactivity.



The aforementioned processing times were calculated for the model implemented on the SpiNNaker platform. Therefore, it should be taken into consideration that 1~ms is the smallest amount of time possible, since it is the minimum timestep of that platform. In case of implementing the proposed model on a different platform, the processing times would change based on the minimum delay allowed on that platform.

After the neurons in \ac{CA}3 spike, a 4~ms rest is required before firing again to prevent the \ac{STDP} rule from being affected by operations other than the one being performed. If they produce any spike in less time, both operations would be mixed, resulting in an unstable or unknown state of the memory. If we take into account this necessary stationary period between operations, we have to distinguish two different temporal concepts in the operations: 

\begin{itemize}
    \item Operation time: time needed to perform the operation, from the instant of time when the information is received at the input until the output is completely stable (after the CA3 inactivity period).
    
    \item Interoperation time: time elapsed from the start of one operation until the instant of time when the next operation can be performed, even if the previous operation has not finished yet.
\end{itemize}

If there were no time constraints, an operation could start right in the next timestep after the previous one; however, as was previously mentioned, \ac{CA}3 introduces some time constraint between operations. Consequently, these temporal concepts will be different for reading (recalling) and writing (learning), as they require receiving the input for a set of consecutive timesteps to perform the operation. For both operations, the following times are detailed:

\begin{itemize}
    \item Operation time (recall): a total of 11~ms are needed. This value is obtained from the sum of the following: 1~ms delay on the INPUT-DG synapses, 2~ms for the recoding of the cue in DG, 1~ms delay from DG to CA3cue, 1~ms for the activation of the CA3 module (CA3cue-CA3cont), 1~ms for the activation of the PC neurons in CA3cont that complete the memory content from the cue, 1~ms delay on the CA3cue-CA1 synapses, 1~ms for recoding the cue in CA1, 1~ms delay on the CA1-OUTPUT synapses, and 2~ms of necessary inactivity in CA3 (it would not be 4~ms, since 2~ms have already elapsed since the last activation of CA3). Therefore, if the cue arrives at ms~1, the result of the recall operation can be seen at ms~9, at ms~11 CA3 will finish stabilising and at ms~12 CA3 can be activated again as a result of another operation.
    
    \item Interoperation time (recall): a total of 6~ms are needed. After 6~ms of starting a recall operation, any other operation can be executed, since the first activation of CA3 as a consequence of the new operation will occur after 4~ms (1~ms INPUT-DG + 2~ms DG + 1~ms DG-CA3). Thus, the activity of the next operation will arrive at CA3 just at the same time as CA3 has stabilised from the previous recall operation. If the recall operation starts at ms~1, the next operation can be started at ms~7 since the first activation of CA3 as a consequence of the next operation will be at ms~12, and the result of the recall will be at ms~9. 
    
    \item Operation time (learning): a total of 12~ms are needed, calculated from the following sum (similar to that of the recall operation): 1~ms INPUT-DG + 2~ms DG + 1~ms DG-CA3 + 3~ms CA3 (due to the need to receive the input for three consecutive timesteps) + 1~ms CA3-CA1 + 1~ms CA1 + 1~ms CA1-OUTPUT + 2~ms of inactivity in CA3. Thus, if a memory arrives at ms~1, it will be stored in CA3 at ms~7, at ms~12 CA3 will finish stabilising and, at ms~13, CA3 can be activated again as a consequence of another operation.
    
    \item Interoperation time (learning): 7~ms in total. It follows the same logic as for the recall operation. If the learning operation starts at ms~1, the next operation can be started at ms~8 (just after the storage of the memory in CA3 at ms~7), since the first activation of CA3 as a consequence of the next operation will be at ms~13.
\end{itemize}

This functionality and temporality was achieved thanks to the configuration of neuron parameters shown in Tab.~\ref{tab:neuronparams}, obtained by means of a grid search algorithm \cite{pontes2016design}. Among all these, low values of $\tau_{synExc}$, $\tau_{synInh}$, $c_{m}$ and $\tau_{m}$ are noteworthy, allowing for fast cycles of membrane potential stabilisation and integration of input spikes. A $\tau_{refrac}$ of 1~ms in CA3 was also used, which is the minimum value necessary for the memory not to oscillate uncontrollably.

\subsection{Summary of resources and timing constraints}
\label{subsec:hippocampus_resources}

\begin{table*}[]
\centering
\caption{Number of neurons and synapses of the different layers of the proposed model. For each layer, the size of the input and output interfaces are also specified. Since input and output layers do not belong to the model itself, only the size of the interface between them and the model is detailed.}
\label{tb:layer}
\resizebox{\textwidth}{!}{%
\begin{tabular}{|l|c|c|c|c|}
\hline
\textbf{Layer}  & \multicolumn{1}{c|}{\textbf{Input neurons interface}} & \multicolumn{1}{c|}{\textbf{Total internal neurons}} & \multicolumn{1}{c|}{\textbf{Total internal synapses}} & \multicolumn{1}{c|}{\textbf{Output neurons interface}} \\ \hline
\textbf{INPUT}  & -                                           & -    & -                                           & $\lceil\log_{2}(cueSize)\rceil + contSize$                                                     \\ \hline
\textbf{DG}     & $\lceil\log_{2}(cueSize)\rceil$                                      & $2^{\log_{2} cueSize + 1} + \log_{2} cueSize + 2$                 & $2^{\log_{2} cueSize} * (2*\log_{2} cueSize + 1) + 3*\log_{2} cueSize + 2$  & cueSize     \\ \hline
\textbf{CA3cue} & cueSize                                     & cueSize                                     & 0       & cueSize                                               \\ \hline
\textbf{CA3cont} & contSize                                      & contSize                                     & 0         & contSize                                            \\ \hline
\textbf{CA1}    & cueSize              & $\lceil\log_{2} cueSize\rceil$                                      & $\sum_{i=2}^{n} ones(bin(i))$ *             & $\lceil\log_{2}(cueSize)\rceil$                                  \\ \hline
\textbf{OUTPUT} & $\lceil\log_{2}(cueSize)\rceil + contSize$   & -                                            & -                                           & -                                                     \\ \hline
\multicolumn{5}{l}{*\textbf{The term \textit{ones(bin(i))} represents the number of 1's in the binary representation of number \textit{i}.}} 
\end{tabular}%
}
\end{table*}

The proposed hippocampal memory model is made up of a set of layers whose required resources can be seen in Table~\ref{tb:layer}. The table shows the number of neurons and internal synapses, as well as the size of the input and output of each layer that makes up the model. In the case of the INPUT and OUTPUT layers, as they are external to the network, only the size of the input and output interfaces is specified.

Summing up the neurons needed per layer, the total number of neurons required by the memory model is $2^{\log_{2} cueSize + 1} + \log_{2} cueSize$ + cueSize + contSize + $\lceil\log_{2} cueSize\rceil + 2$ neurons. On the other hand, if we look at the amount of internal synapses for each layer (Table~\ref{tb:layer}) and the interconnections between them (Section~\ref{subsubsec:connections}), a total of $2^{\log_{2} cueSize} * (2*\log_{2} cueSize + 1) + 3*\log_{2} cueSize$ + $2*cueSize$ + $2*contSize$ + $2*\lceil\log_{2}(cueSize)\rceil$ + $\sum_{i=2}^{n} ones(bin(i)) + 2$ static synapses are needed, plus $cueSize*contSize$ dynamic ones using STDP.

As for the temporal aspect, the memory can perform a new operation every 6~ms if the previous one was a recall or 7~ms if it was a learning operation. These times cannot be shorter because of the need for a 4~ms resting period in CA3 between input activities of different operations to prevent the STDP rule from acting in an undesirable way. In addition, the result of the recall operation, i.e., the complete recalled memory, will be obtained after 8~ms after the cue has been inputted to the memory. More details can be found in Section~\ref{subsubsec:implemented_functionality}.

\section{Experimentation and results}
\label{sec:experiments_and_results}

To demonstrate the functioning of bio-inspired hippocampal memory, a set of incremental experiments were developed. The purpose of these was to first prove the basic operations of learning, recalling and forgetting in detail. On the basis of these three operations, more complex experiments were performed with combinations of these operations, random accesses and a stress test. 

For these experiments, a network capable of storing a total of 5 memories simultaneously (cueSize) and with a memory size of 13 \ac{PC} neurons (contSize+cueSize) was used. Therefore, the activity of the first 3 neurons corresponds to the cue of the memory and the activity of the other 10 neurons corresponds to the content of the memory. Internally, the representation of the cue will need a size of 5 neurons to be able to store the 5 memory cues simultaneously once encoded in one-hot.

For the last experiment, a much larger memory was used in order to evaluate the performance and prove the timing and operation constraints. A model capable of simultaneously storing 64 memories of a size of 38 neurons each was used. In this case, the first 6 neurons encoded the cue, whereas the other 32 neurons encoded the content of the memory, and, internally, the representation of the cue will need 64 neurons to deal with the one-hot encoding of the cues.

\subsection{Learning}
\label{subsec:learning}

\begin{figure*}[!t]
    \centering
    \includegraphics[width=0.75\textwidth]{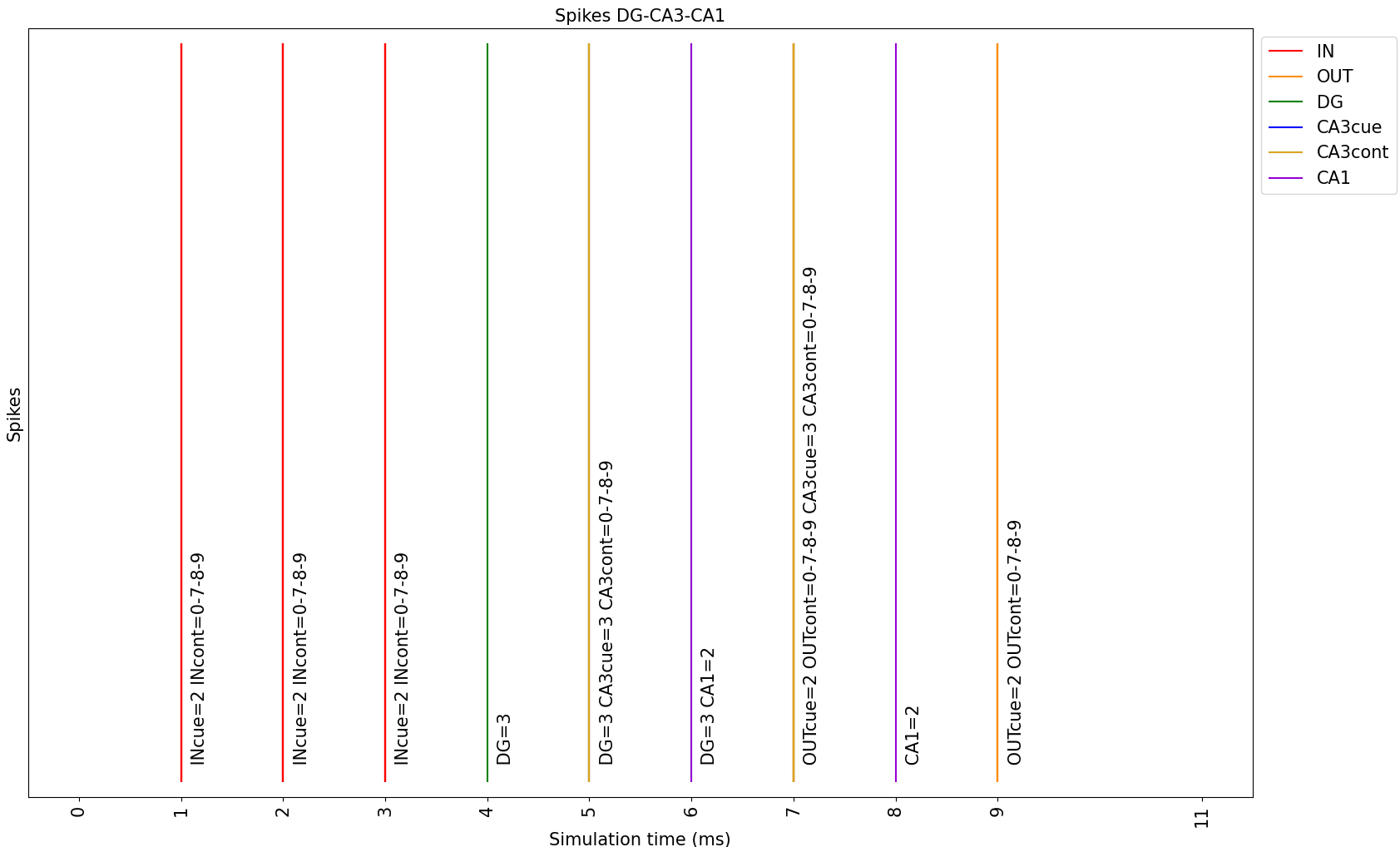}
    \caption{Internal and external spike activity of the proposed hippocampal memory model during a learning operation. The simulation starts with the model receiving the input of the cue and the content of the memory to be stored, which then activates the corresponding modules of the memory. After the processing time needed, the model outputs the same activity that was used as input, after being stored by means of the STDP rule.}
    \label{fig:test1}
\end{figure*}

In the first experiment, a single learning operation is performed. Fig.~\ref{fig:test1} shows the spiking activity of all the different layers of the  memory model over time. The presence of spikes is denoted as a vertical bar whose associated mark indicates which layers and which neurons in those layers have spiked at that particular time instant.

In this experiment, the learning operation attempts to store a memory formed by the pattern of input activity corresponding to the activation of neurons 2, 3, 10, 11, 12 and 13, simultaneously. Splitting this information into cue and content, the cue is determined by the activation of neuron 2, and the content is formed by the activation of neurons 0, 7, 8 and 9.

The operation starts at ms~1 and requires holding the input for three consecutive timesteps (3~ms), hence the activation of the input layer at ms~1, ms~2 and ms~3. Then, the part of the input activity corresponding to the cue is sent to DG, in which it is recoded.

As only the cue is recoded and is represented by the activation of neuron 2, the corresponding one-hot coding would be given by the activation of neuron 3 in DG. In binary, the spatial representation of the original input cue would be 4 (100, which corresponds to the activation of neuron 2); therefore, in one-hot encoding, it would be represented as 01000 or, in other words, as the activation of neuron 3 (starting from 0).

DG takes 2~ms to process the input and requires an additional 1~ms for receiving the input from INPUT, thus the input activity corresponding to ms~1, ms~2 and ms~3 will have an output activity in DG at ms~4, ms~5 and ms~6, respectively. The recoded cue from DG and the rest of the memory content from the input will arrive simultaneously at CA3. 

DG output activity at ms~4 and ms~6 will cause the output activity in CA3 at ms~5 and ms~7, respectively. Since the neurons in CA3 have a refractory period of 1~ms, they cannot generate spikes in consecutive timesteps. Therefore, the DG output activity at ms~5 will not have a corresponding CA3 output activity at ms~6, as CA3 will be in refractory period as a consequence of its activation at ms~5 and will not integrate these new spikes.

The activity of CA3 at ms~5 and ms~7 corresponds to the encoded cue and the content of the memory. The activity related to the content arrives at CA3cont completely unchanged, since it was inputted to the model, and therefore, follows the same activation pattern (0, 7, 8 and 9 in this case). In CA3cue, on the other hand, the cue arrives encoded as the activation of a single neuron (neuron 3). By activating these corresponding neurons in both CA3 populations at the same time twice during this process, a double activation of the STDP learning rule is achieved. Two activations are enough for the weights to saturate to the maximum possible value and, thus, to store the memory correctly in the synapses. These synapses would be those that connect neuron 3 of CA3cue with neurons 0, 7, 8 and 9 of CA3cont.

Although the storage of the memory finishes at ms~7, the CA3 activity continues propagating through the network until reaching the output. The cue, i.e., the activity related to CA3cue, is passed to CA1 for being recoded. The recoding process back to the original input activity takes 1~ms, and thus the activation of CA3cue at ms~5 and ms~7 makes CA1 generate outputs at ms~6 and ms~8, respectively. In this case, CA3cue neuron 3 generates a spike, representing 01000 in one-hot, which is encoded as 100 (activation of CA1's output neuron 2) in CA1. 

Finally, the complete learned memory (the same as the memory that arrived at the input) is outputted. The activity related to the cue arrives from CA1 at the same time as the activity representing the content of the memory arriving from CA3cont. Thus, the input memory can be seen in OUTPUT at ms~7 and ms~9 as a consequence of the output activity of CA1 (which imposes the temporal order with which the output of CA3cont must be synchronised) at ms~6 and ms~8.

The purpose of this experiment was to verify that the temporal and sequential aspects of the produced spikes matched the expected behaviour, and that the weight changes taking place in the dynamic synapses as a consequence of the activation of the STDP rule in CA3 saturated positively only in the corresponding cases.

\subsection{Learning and recall}
\label{subsec:learning_and_recall}

\begin{figure*}[!t]
    \centering
    \includegraphics[width=0.75\textwidth]{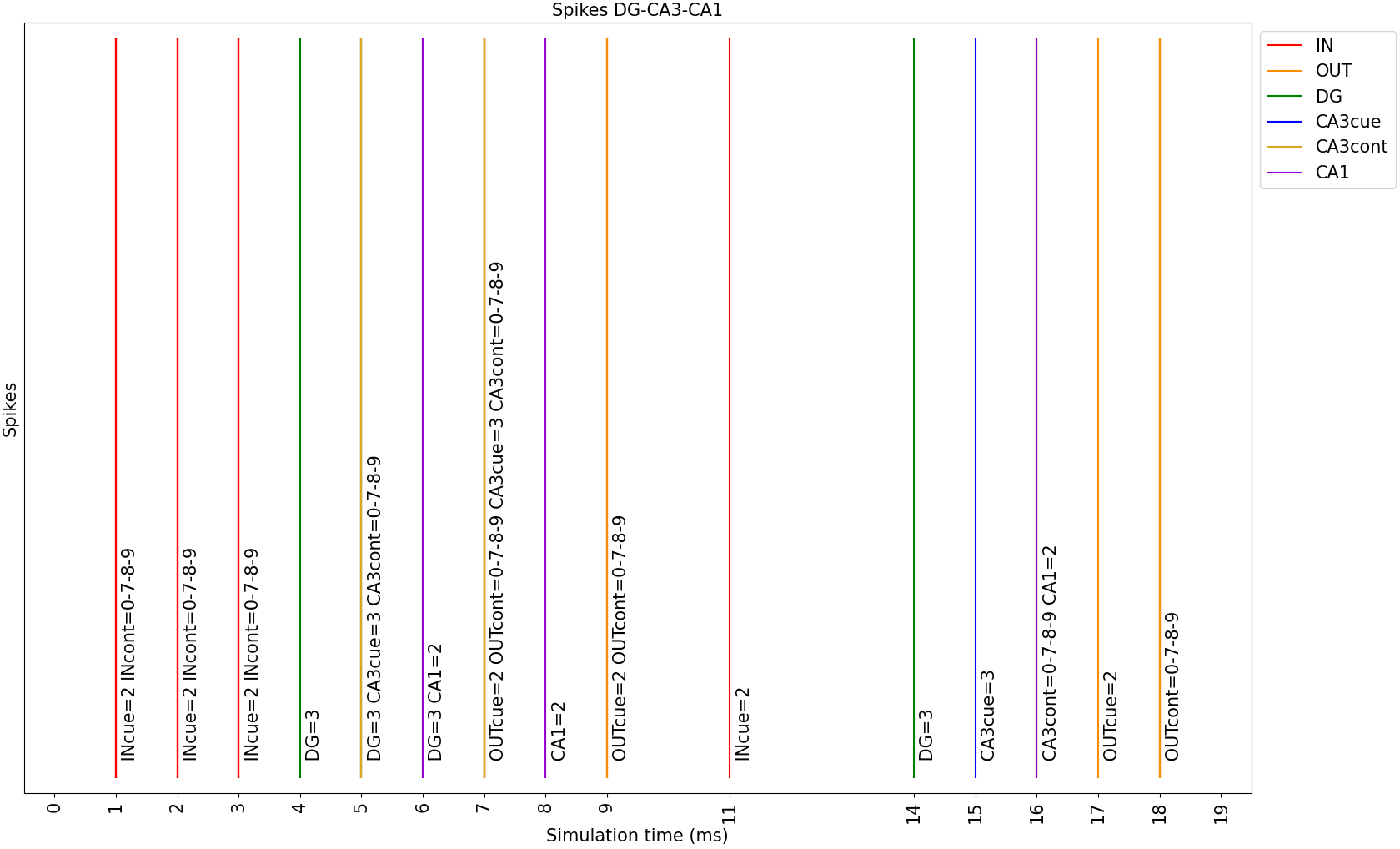}
    \caption{Internal and external spike activity of the model during a sequence of learning and recalling operations. During the first half of the simulation, the learning phase of the memory can be observed, while the second half (beginning at ms~11) corresponds to the recalling process.}
    \label{fig:test2}
\end{figure*}

This experiment was based on the previous one and extends it by adding a recall operation after the learning operation. Two main goals were pursued with this experiment: to check the correct functioning of the recall operation, i.e., whether the memory is capable of recalling a previously stored memory and to analyse whether the learning operation is capable of correctly storing the memory. The model used in this experiment is the same as in the previous experiment. The cue consists in the activation of neuron 2 (and the content of the memory) by the activation of neurons 0, 7, 8 and 9.

Fig.~\ref{fig:test2} shows the spiking activity of all the different layers of the model (including input and output) over time. The learning operation follows the same sequence of spikes as discussed in the previous experiment. Following the temporal constraints specified in section \ref{subsubsec:implemented_functionality}, the learning operation starts at ms~1 and the memory is stored at ms~7. However, due to the resting time needed between operations in CA3, the recall operation does not start until ms~8. For the sake of simplicity, in this experiment, the recall operation was set to start at ms~11 instead, so that the spiking activity in the figure can be seen more clearly.

To run the recall operation, the only element needed for the model is to receive a cue for just a single millisecond in the input. At ms~11, the input cue (the activation of neuron 2 in this case) is provided to the model. Following the same temporal and logical sequence as in the learning operation, this activity reaches DG and recodes the cue 3~ms later, resulting in the activation of neuron 3 at ms~14. This signal propagates to CA3, where CA3cue neuron 3 will be activated at ms~15.

Thanks to the previous learning, the activation of the cue (neuron 3 of CA3cue) triggers the recall of the content associated (neurons 0, 7, 8, and 9 of CA3cont) at the next instant (ms~16). From here, the activity related to the cue is recoded as it passes through CA1, leaving the memory at ms~16 and generating the activation of the corresponding neuron in the output population at ms~17. The rest of the recalled content of the memory (CA3cont activity) is outputted with its respective delay at ms~17, activating the population connected to the output at ms~18.

This proves the correct functioning of the learning and recall operations, since a complete memory has been first stored and then retrieved correctly.

\subsection{Learning, forgetting and relearning}
\label{subsec:learning_forgetting_relearning}

\begin{figure*}[!t]
    \centering
    \includegraphics[width=0.75\textwidth]{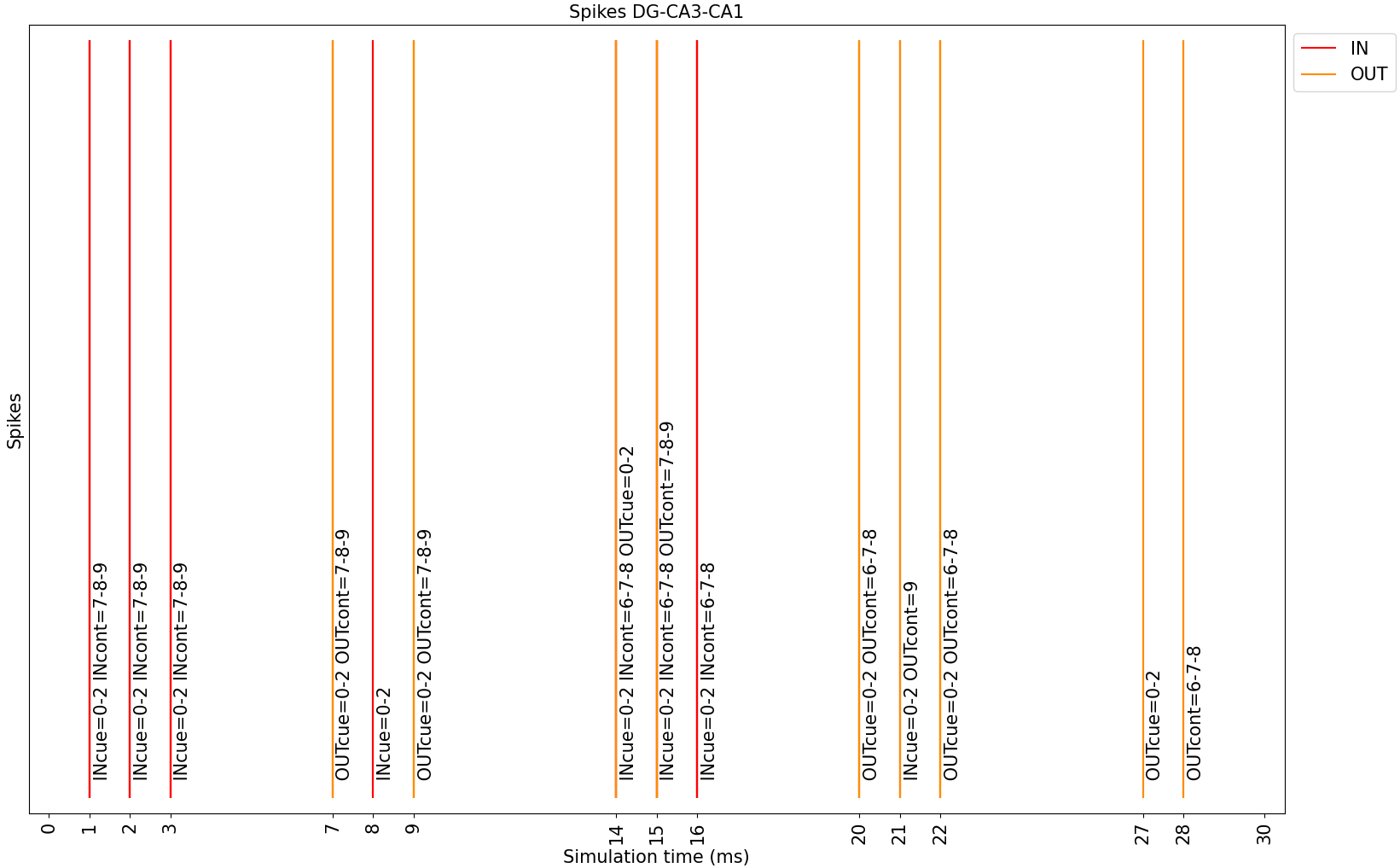}
    \caption{Input and output spike activity of the hippocampal-based memory during the sequence of learning, recall, forgetting-relearning and recall operations with the same cue. The learning and recall of the first pattern can be observed in the first half of the simulation. The second half of the simulations correspond to the forgetting of the first pattern and learning of the second pattern with the same cue as the first one.}
    \label{fig:test3}
\end{figure*}


As was already discussed in Section~\ref{subsubsec:implemented_functionality}, the forgetting of a memory is not an operation as such, but a functionality that is automatically derived from learning different contents that share the same cue.

This experiment builds on the previous one, where the learning and recalling of a memory is carried out, and extends it with another learning-recalling pair of operations where a different memory that shares the same cue as the first one is used. In this way, the memory has to forget the first memory in order to learn the second one.

The first memory consists of a cue where neurons 0 and 2 are activated, and the content is formed by the activation of neurons 7, 8 and 9. The second memory has a neuron activation pattern identical to the previous one for the cue (neurons 0 and 2), while for the content is formed by the activation of neurons 6, 7 and 8. As can be observed, both memories have not only the same cue but also part of the same content (neurons 7 and 8). These patterns were selected in order to prove that the content of different memories do not have to be fully orthogonal in order to be stored correctly.

Fig.~\ref{fig:test3} shows the spiking activity diagram over time. For the sake of simplicity, only the input and output layers are shown, for two main reasons. Firstly, the more operations, the harder it is to visually understand the activity of all layers (since the activity of different layers and operations would be overlapped). Secondly, the two main memory operations (learning and recall) have already been explained in detail in previous experiments, thus a detailed representation is not necessary.

In the first half of the simulation, a learning and recall operation is performed for the first memory. The learning input activity starts at ms~1 and the learning output activity ends at ms~9. In the case of the recall, unlike in the previous experiment, it was performed as soon as possible, hence starting at ms~8 and ending (complete recall is obtained) at ms~15. This recall operation started even before the previous one ended; however, as the flow of activity travels through different layers and the minimum resting time of CA3 is respected, there is no conflict between the two operations (see Section~\ref{subsubsec:implemented_functionality}).

These operations have a flow of activity that is identical to that explained in previous experiments but applied to the new memory. It should be noted that, in this case, the cue consists in the activation of neurons 0 and 2, which, using the spatial representation in binary, would correspond to the value 5 and, therefore, to 10000 in one-hot (activation of neuron 4).

In the second half of the simulation, another pair of learning-recall operations is performed, but with the second memory. The learning operation takes place from ms~14 to ms~22 and the recall operation from ms~21 to ms~28. 

In this second half, the memory that is wanted to be stored has the same cue as the one that was previously stored in the first half. When the second memory (cue and content) is used as input in the learning operation, it automatically triggers a recall operation for the first one (because of the cue). Therefore, during the 3~ms of input activity (ms~14-16) needed for the learning, the first and third ms activate CA3 with the second memory (ms~20 and 22) and, in the second one (ms~21), the activation of the content of the first memory in CA3 is produced due to the involuntary recall operation. Specifically, the memory only activates the neurons of the first memory that are not present in the second memory, since the neurons that coincide with the second memory will be in the refractory period due to the activation in the previous ms.

This effect has two consequences: those synapses that store the second memory will be reinforced, while those that store the first memory (and that are not present in the second memory) will saturate their weight to 0, forgetting the learned pattern. Even if the activity of the internal components of CA3 is not present in Fig.~\ref{fig:test3}, the aforementioned effects can be observed in the output activity of the second learning operation. At ms~20 and ms~22, the whole second memory is returned, and, at ms~21, the part of the content of the first memory that is not present in the second memory will thus be forgotten (neuron 9). Finally, the second recall operation succeeds in correctly retrieving the complete second memory without any trace of the first memory, indicating that the forgetting operation has been performed correctly.

\subsection{Benchmarks}
\label{subsec:tests}

\begin{figure*}[!t]
    \centering
    \includegraphics[width=0.8\textwidth]{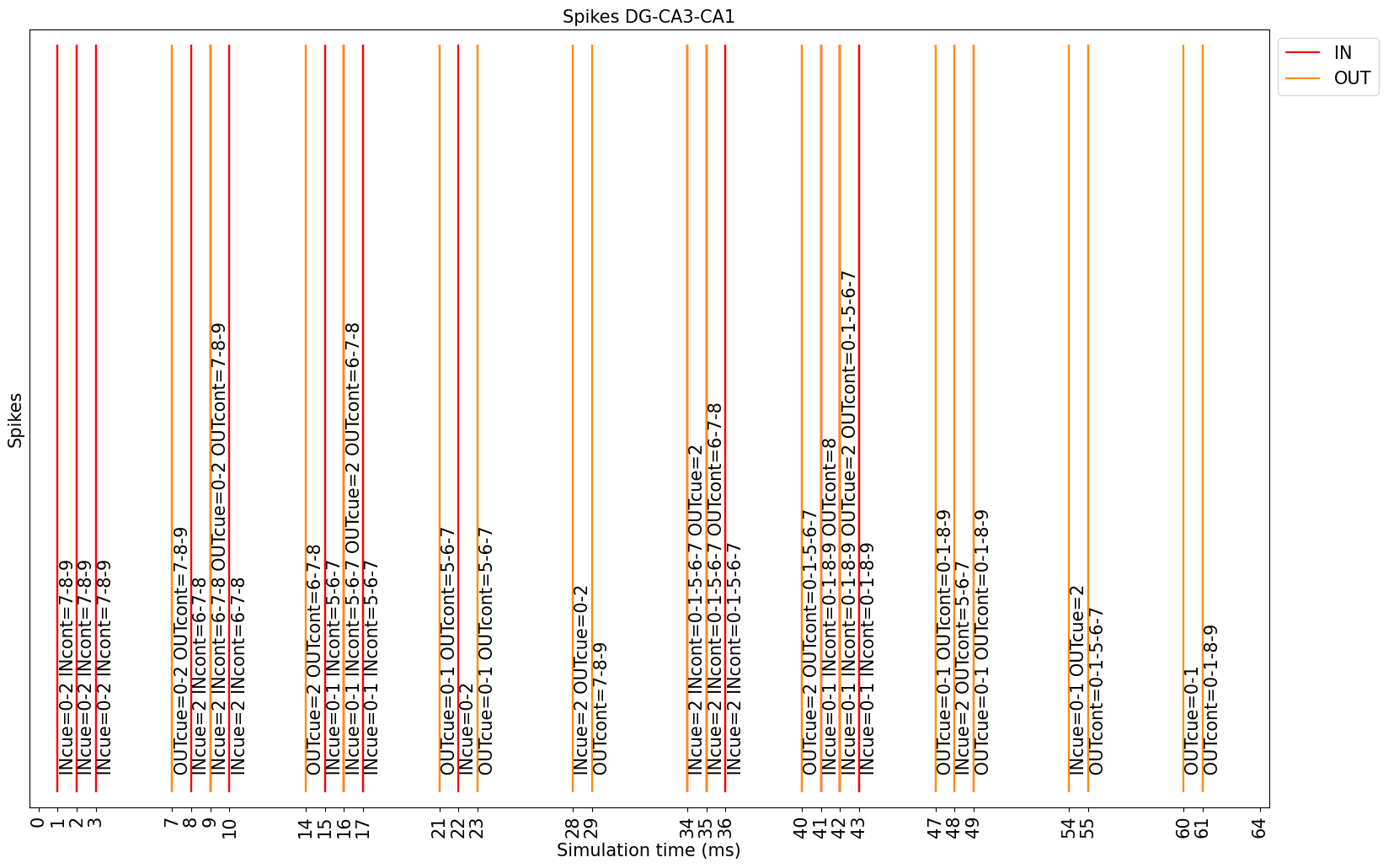}
    \caption{Input/output spike activity of the hippocampus memory model in a test based on the combination of learning, recall and forgetting-relearning operations.}
    \label{fig:test4}
\end{figure*}

After checking the correct functioning of all the implemented memory operations, a set of more complex tests were evaluated to verify the robustness of the memory. Three different tests were carried out:

\begin{itemize}
    \item Combined operations sequence test: consists of a set of learning, recalling and forgetting operations combined, following a regular case of memory use, as can be seen in Fig~\ref{fig:test4}. A total of 5 learning operations were performed (two of which involve forgetting a previous memory) and 4 recall operations, with a total duration of 64~ms.
    
    \item Random learning and recall test: it is a random access test both in terms of operations and in terms of the cue and content of the memories. A total of 100 completely random operations were executed, which had a duration of 657~ms. Among these operations, 46 corresponded to learning operations and 54 to recall operations.
    
    \item Stress memory test: a much larger memory capacity than in previous tests was used (able to store a total of 64 memories of 38 neurons each) and several memory sweeps with different operations were carried out to test the memory. By sweeps we refer to the fact that the operations were performed for each possible cue that could be stored in memory. This test is based on the MemTest86 algorithm\footnote{“Memtest86 - The standard for memory diagnostics,” \url{http://www.memtest86.com/}.}, which is a well-known memory testing software for x86 and ARM computers. The sequence of operations performed consisted of pairs of learning-recalling operations with memories using all the possible cues. The first learning sweep stored the spatial binary encoding of the cue as the content of the memory, while the other learning sweeps carried out the learning of the complementary of that content. In this way, for each pair of learning sweeps, the original memory was retrieved again. The whole test consisted of 384 operations and lasted 2497~ms. Among these, 192 were learning operations (128 of which included forgetting the previous memory) and 192 recall operations, i.e., 3 complete memory sweeps.
\end{itemize}

The results of the first test can be seen in Fig.~\ref{fig:test4}. However, the other two reached the order of seconds of simulation, and a massive amount of activity was generated, which is difficult to represent graphically. In all of them, the memory was able to correctly perform all the learning and recalling operations performed, with some observations regarding the stress test that are discussed in Section~\ref{sec:discussion}.

\section{Discussion}
\label{sec:discussion}

The results of the experiments performed in Section~\ref{sec:experiments_and_results} demonstrated the correct functioning of the basic memory access operations: learning, recalling and forgetting. Moreover, the different tests that were carried out were passed as expected. Therefore, the functioning of the proposed spike-based memory model was proved, not only in individual and independent operations, but also in real-case scenarios, random accesses and even under conditions of great stress with large amounts of accesses. These experiments served to validate the bio-inspired theoretical bases on which the hippocampal bio-inspired memory model is based, to confirm its correct functioning and, in addition, to show the flow of internal activity and input/output of the spiking network model.

The proposed model passed the tests correctly, although with some limitations, particularly during the stress test. After performing many consecutive memory operations between one access and the next to a specific memory, either between learning and recall operations or between different recall operations, the memory is eventually forgotten. In other words, for each memory operation that does not involve a stored memory, the synapses that store that memory are degraded, and thus the memory itself is degraded.

This is due to the way in which the \ac{STDP} learning mechanism is implemented in SpiNNaker. After a period of time between presynaptic and postsynaptic spikes longer than the spike integration period of the dynamic synapses ($\tau_{minus}$ and $\tau_{plus}$), the increase or decrease (depending on the temporal causality of these spikes) of the weight at these synapses tends to 0. In many implementations, the weight variation is usually truncated to 0 after this period has passed. However, in SpiNNaker, a residual value greater than 0 is maintained. This means that, for each operation performed in which part of the content of an already stored memory participates and it is not the same memory involved in the operation, a small decrease in the weights of its synapses is produced and, therefore, the content is slightly degraded. This memory will be slightly degraded for each operation related to a different memory that shares part of the content, until it is completely forgotten in memory if it is not accessed again first.

Although this characteristic of the model could be considered a limitation from a functional point of view, there is evidence linking it to natural processes found in biology \cite{rolls2021brain}. The hippocampus stores all the information that arrives through the sensory flows at each instant in the form of memories, although not all this information is necessary; it requires filtering in order to retain the relevant information. The degree of relevance of memories is measured by the number of times they are recalled and, in general, accessed. Those memories that are accessed will be reinforced, and those that are not accessed will be degraded over time. At some point, the former will be integrated and held in memory longer, as they are considered relevant and useful, while the latter will be completely forgotten, as they have not been needed for a long time \cite{rolls2021brain}.

This is also the case in the proposed bio-inspired model of the hippocampus, since those memories that are recalled are retained in memory and those that are not will be degraded and eventually forgotten. This degradation of the content of the memory could be seen as a rather interesting feature from a functional point of view that could be useful in many applications, as it indirectly denotes the relevance of memory content in an automatic way.

As in the biological model, the proposed model is able to store all kinds of information as memories within the memory, regardless of its format and source, without the need to give a prior interpretation of the content. Moreover, the biological model shows an oscillatory and unstable activity and can recall an entire memory from any part of its content, whereas the proposed model shows a regulated and deterministic activity that can recall the memory from the content that is initially considered as a clue.

By opting for a regulated activity approach, the activity in the network occurs only when necessary, thus increasing energy efficiency. Furthermore, although a memory can be theoretically retrieved from any part of its content, that part must be sufficiently identifiable to only recall that single memory; otherwise, a combination of several memories would be recalled at the same time. This suggests that not all parts of the memory are used in its recalling, but only those that allow them to be uniquely identified.

Another aspect with some ambiguity is the coding achieved in DG and CA1. It is known that, in their biological counterparts, the coding they perform increases the sparsity of the memories to increase storage capacity, but not the degree of sparsity achieved. For the proposed model, it was decided to implement the highest possible sparsity, i.e., converting the information to one-hot, but only to the part that really requires it: the cue. Therefore, consistency with the biological model is maintained in these regions.


Regarding the models proposed in the literature, the model described in this paper is compared with those mentioned in Section~\ref{sec:introduction} that deal with a spike-based bio-inspired hippocampus memory model. Unlike the models proposed in \cite{tan2011associative, tan2013hippocampus}, the one presented in this paper requires only a single CA3 network to learn and recall correctly. Our model was tested with a capacity of up to 64 patterns of 38 neurons of activity each (allowing for higher, but not evaluated, capacities), as opposed to the sequence of 11 patterns of 1 neuron each proposed in the aforementioned works. The hippocampal memory model presented in this paper was designed entirely with \acp{SNN}, while the one proposed in \cite{zhang2016hmsnn} is partially spike-based.

Finally, the memory models proposed in \cite{casanueva2022spike} are not able to work correctly with non-orthogonal patterns and cannot combine phases of learning and recalling of memories, which are characteristics of the model presented in this work. Moreover, in the mentioned work, only the CA3 network with oscillating behaviour was implemented, whereas, in the present work, all hippocampal layers are modelled, although with a deterministic and regulated behaviour. Overall, this paper presents a complete spike-based bio-inspired hippocampus memory model that is much more functional and applicable to real cases than those proposed in the literature.

\section{Conclusions}
\label{sec:conclusions}

In this work, a fully functional bio-inspired spike-based hippocampal memory model was proposed and designed using \acp{SNN} implemented on the SpiNNaker hardware platform. The architecture of the proposed model is fully parameterised, thus it can be implemented with the desired amount of storable memories and different maximum memory sizes. Specifically, it has been shown to perform well for memories with storage capacities of up to 64 memories in parallel with 38 neurons of activity each.

Through a set of incremental experiments, the operation of basic memory access operations was demonstrated: learning, recalling and forgetting of both orthogonal and non orthogonal memories (memories that have some neural activity in common). This first set of experiments served as a basis for analysing the internal flow of the model's activity, from the time when neural activity enters as a total or partial memory until the model gives an output response to this input. The experimentation includes a series of tests to check the capabilities of the memory in different situations, such as a common use case, completely random accesses and huge amounts of operations reaching a maximum stress scenario for the memory.

These experiments were successfully carried out, demonstrating the model's ability to learn input memories, recall them using a cue as input and even forget them if different memories with the same cue are learned over time. However, due to the implementation of the STDP mechanism in SpiNNaker, an additional behaviour also observed in biology was found. For each memory access, the accessed memory will be strengthened, while all other stored memories that share part of the content with it will be slightly degraded; therefore, over the course of several operations, if these memories are not accessed, they will be forgotten. Indirectly, the memory model is able to determine the relevance of the stored content automatically, maintaining those memories it considers relevant longer and forgetting those that it does not. Relevance is determined by the number of accesses to the memory and their temporal proximity.

Furthermore, the proposed memory model was compared not only with its biological counterpart, but also with other computational models that can be found in the literature. This work presents the first hardware implementation of a fully functional bio-inspired spike-based hippocampus memory model, paving the road for the development of future more complex neuromorphic systems.


The limitations of the model, the characteristics of the biological model of the hippocampus on which the proposed memory is based and the intrinsic capabilities of SNNs open a research line for future work exploring these aspects. To test the effectiveness of the proposed model, it could be implemented on other hardware platforms, such as Loihi and, through a series of tests, it would be interesting to verify whether the integrity of its operation is still maintained, considering that this platform has a lower timestep (0.19 $\mu$s). Taking features present in the biological model, it would be possible to create output connections that connect to the input, via an intermediate module, to produce a feedback memory that is capable of storing and replaying sequences, which is a capability observed in the hippocampus. The hippocampus is able to recall a memory from any part of it that is representative; in the presented model, it can recall any memory from a fragment of it (the cue), not the whole memory, thus it would be interesting to explore the addition of dynamic STDP synapses connecting CA3cont to CA3cue, allowing for the recall of the memory cue from the content. Overall, there are many possibilities for applications with a memory module in the spike-based paradigm.

The source code regarding the implementation of the fully functional hippocampal bio-inspired memory with \acp{SNN} on SpiNNaker is available on GitHub\footnote{\url{https://github.com/dancasmor/A-bio-inspired-implementation-of-a-sparse-learning-spike-based-hippocampus-memory-model}}.

\bibliographystyle{IEEEtran}
\bibliography{bibliography}






\vspace{-30pt}

\begin{IEEEbiographynophoto}{Daniel Casanueva-Morato}
received the B.S. degree in computer engineering from University of Seville, Spain, in 2020 and the M.S. degree in computer engineering and networks from University of Granada, Spain, in 2021. Currently, he is a Ph.D. student in the Departmenf of Computer Architecture and Technology at the University of Seville, supported by a "Formación de Personal Universitario" Scholarship from the Spanish Ministry of Education, Culture and Sport. His research focuses on computational neuromorphic architectures for modeling brain regions. His research interests include neuromorphic engineering, spiking neural networks, brain-computer interfaces and embedded systems.
\end{IEEEbiographynophoto}

\vspace{-30pt}

\begin{IEEEbiographynophoto}{Alvaro Ayuso-Martinez}
received the B.S. degree in computer engineering from University of Seville, Spain, in 2020 and the M.S. degree in computer engineering and networks from University of Granada, Spain, in 2021. Currently, he is a PhD student in the Robotics and Technology of Computers Lab. in the Department of Computer Architecture and Technology at Universidad de Sevilla. His research focuses on neuromorphic engineering, more specifically on spiking neural network architectures, being part of the MIND-ROB research project. His research interests include neuromorphic engineering, spiking neural networks, real-time audio processing and embedded systems.
\end{IEEEbiographynophoto}

\vspace{-30pt}

\begin{IEEEbiographynophoto}{Juan P. Dominguez-Morales}
received the B.S. degree in computer engineering, the M.S. degree in computer engineering and networks, and the Ph.D. degree in computer engineering (specializing in neuromorphic audio processing and spiking neural networks) from the University of Seville, in 2014, 2015 and 2018, respectively. His Ph.D. was granted with a research grant from the Spanish Ministry of Education and Science. Since January 2019, he has been working as Assistant Professor in the same university. His research interests include neuromorphic engineering, spiking neural networks, audio processing and deep learning.
\end{IEEEbiographynophoto}

\vspace{-30pt}

\begin{IEEEbiographynophoto}{Angel Jimenez-Fernandez}
received the B.S. Degree in Computer Engineering in 2005, the M.S. Degree in Industrial Computer Science in 2007 and the Ph.D. in Neuromorphic Engineering in 2010 from the University of Seville, Sevilla, Spain. In October 2007 he became Assistant Professor at the department of Computer Architecture and Technology of the University of Seville. His research interests include neuromorphic engineering applied to robotics, real-time spikes signal processing, neuromorphic sensors, FPGA design, embedded intelligent systems development, and biomedical robotics.
\end{IEEEbiographynophoto}

\vspace{-30pt}

\begin{IEEEbiographynophoto}{Gabriel Jimenez-Moreno}
received the M.S. Degree in Physics (electronics) and the Ph.D. Degree from the University of Seville (Seville, Spain), in 1987 and 1992, respectively. He was granted a Fellowship from the Spanish Science and Technology Commission (CICYT). Currently, he is a Full Professor at the University of Seville. From 1996 until 1998, he was Vice-Dean of the E.T.S. Ingenieria Informatica, University of Seville, where he participated in the creation of the Department of Computer Architecture. He is the author of several papers on robotics, neuromorphic engineering, and computer architecture, and has also directed many research projects on neuromorphic systems. His research interests include neural networks, vision processing systems, embedded systems, computer interfaces, and computer architectures.
\end{IEEEbiographynophoto}

\vfill

\end{document}